# Epistemology of Generative AI: The Geometry of Knowing

Ilya Levin

Holon Institute of Technology, Holon, Israel

[levini@hit.ac.il](mailto:levini@hit.ac.il)

**Abstract**

Generative AI presents an unprecedented challenge to our understanding of knowledge and its production. Unlike previous technological transformations, where engineering understanding preceded or accompanied deployment, generative AI operates through mechanisms whose epistemic character remains obscure—and without such understanding, its responsible integration into science, education, and institutional life cannot proceed on a principled basis. This paper argues that the missing account must begin with a paradigmatic break that has not yet received adequate philosophical attention. In the Turing–Shannon–von Neumann tradition, information enters the machine as encoded binary vectors and semantics remains external to the process. Neural network architectures rupture this regime: symbolic input is instantly projected into a high-dimensional space where coordinates correspond to semantic parameters, transforming binary code into a position in a geometric space of meanings. It is this space that constitutes the active epistemic condition shaping generative production. Drawing on four structural properties of high-dimensional geometry—concentration of measure, near-orthogonality, exponential directional capacity, and manifold regularity—the paper develops an Indexical Epistemology of High-Dimensional Spaces. Building on Peirce's semiotics and Papert's constructionism, it reconceptualizes generative models as navigators of learned manifolds and proposes *navigational knowledge*—a third mode of knowledge production, distinct from both symbolic reasoning and statistical recombination.

**Keywords:** generative AI, high-dimensional geometry, epistemology, indexical signification, navigational knowledge, manifold hypothesis, Peirce, constructionism

## 1 Introduction

The rapid development of generative AI has produced a remarkable conceptual asymmetry. The systems themselves—GPT-4, Claude, Gemini, Stable Diffusion, and their successors—operate by navigating continuous vector spaces of extraordinary dimensionality. They neither parse logical formulae nor consult symbolic knowledge bases; they activate regions of high-dimensional geometry in response to contextual prompts, producing outputs that are coherent, semantically rich, and sometimes strikingly novel. Yet the theoretical discourse that has grown up around these systems almost entirely ignores the medium in which they operate. Explanations of generativity are sought in architectures (the transformer, the diffusion process), in training regimes (self-supervision, reinforcement learning from human feedback), in data scale (the sheer volume of text or images consumed), or in cognitive analogies (does the system "understand"? does it "reason"?). High-dimensional space itself—the geometric arena in which all of these processes unfold—is treated as a neutral, transparent container: a technical substrate that enables computation but contributes nothing to meaning.

This paper argues that this treatment leaves a crucial dimension unexamined. The geometry of high-dimensional space is not a passive backdrop; it is an active epistemic condition that shapes what generative systems can do, how they produce novelty, and what kind of knowledge their outputs constitute. The structural properties of high-dimensional Euclidean spaces—properties that have no counterpart in the low-dimensional geometries accessible to human perception—create the conditions under which a new form of knowledge production becomes possible. Without attending to these properties, we cannot adequately explain why generative AI works, what it produces, or what its outputs mean.

## 1.1 Current Explanatory Strategies and Their Common Focus

The need for such attention becomes apparent when we survey the current landscape of responses to generative AI. The discourse is wide-ranging and often technically sophisticated, yet it tends to converge on a set of recurring explanatory strategies that, for all their insight, share a common gap: none of them examines the geometry of the representation space as an epistemic condition in its own right.

One prominent strand of this discourse emphasizes the absence of symbolic reasoning. Bender & Koller (2020) argue that language models, however fluent, cannot access meaning because they lack grounding in communicative intent. Bender et al. (2021) sharpen this critique into the *stochastic parrots* thesis: generative systems merely recombine statistical patterns, and any appearance of genuine novelty is illusory. Marcus (2022) catalogues failures of compositional reasoning to support a similar conclusion. These arguments inherit an important intellectual tradition—the symbolic AI tradition's insistence that cognition requires rule-governed manipulation of discrete representations (Newell & Simon, 1976; Fodor & Pylyshyn, 1988)—and they identify real limitations. But in focusing on what is absent—explicit symbolic machinery—these analyses tend to leave open the question of whether epistemic structure might take a different form. The possibility that structure could reside in the geometry of the space rather than in explicit rules remains largely unexplored.

A second strand treats generativity as a purely engineering phenomenon. Kaplan et al. (2020) demonstrate that model performance follows smooth scaling laws: more data, more parameters, more compute yield predictable improvements. Brown et al. (2020) show that few-shot learning emerges as a property of sufficient scale. From this perspective, generative AI is technically impressive but does not appear to require new epistemological categories—it is optimization at scale, continuous with existing computational paradigms. Yet the pragmatist position, for all its technical rigor, leaves a philosophical question unanswered: it describes *how much* performance improves without explaining *why* scaling produces qualitative shifts in expressive capability. The scaling laws are empirically robust, but they are not self-explanatory. Something about the medium in which these models operate must make scaling productive—and that medium is high-dimensional vector space. To observe that performance scales with parameters is to describe a regularity; to explain why it scales is to make a claim about the geometric structure that makes scaling possible.

A third strand oscillates between fascination and alarm. Bubeck et al. (2023) describe GPT-4 as exhibiting "sparks of artificial general intelligence." Popular discourse vacillates between utopian and dystopian framings. This response, while psychologically understandable, tends to obscure what most needs explaining. By attributing to generative systems either too much (quasi-human creativity and understanding) or too little (mere statistical mimicry), it makes it harder to understand generativity on its own terms—as a form of knowledge production that is neither human-like reasoning nor trivial recombination.

What these otherwise divergent positions have in common is that none of them foregrounds the structural medium itself. Whether the focus is on symbolic capacity, on engineering scale, or on cognitive analogy, the high-dimensional vector space in which generative processes unfold tends to be treated as epistemologically transparent—a technical substrate rather than an active condition of knowledge production. This paper proposes that it is anything but transparent. The central thesis is that generative AI cannot be understood without examining the epistemic consequences of high dimensionality—consequences that are not extrapolations of low-dimensional intuition but genuine structural novelties that emerge only when dimensionality crosses certain thresholds (Vershynin, 2018).

To articulate these consequences, this paper develops an *Indexical Epistemology of High-Dimensional Spaces*. This framework draws on Peirce's semiotics to characterize the mode of signification at work in embedding spaces—a mode that is indexical rather than symbolic, grounded in positional relation rather than conventional correspondence. It extends Papert's constructionist epistemology beyond its original symbolic setting, reconceptualizing the high-dimensional manifold as a post-symbolic microworld in which knowledge is enacted through navigation rather than assembled through programming. The resulting account occupies a position that the prevailing explanatory strategies cannot reach: it takes generativity seriously as a form of knowledge production without either anthropomorphizing it or reducing it to statistical accident.

### 1.2 Methodological Position: Structural Epistemology

This paper pursues what might be called a *structural epistemology* of high-dimensional spaces. The term requires clarification, since it marks a deliberate departure from several adjacent methodological traditions.

The paper is not an exercise in the philosophy of mathematics: it does not examine the ontological status of high-dimensional spaces or the foundations of geometric reasoning. Nor is it an exercise in philosophy of mind applied to artificial systems: it makes no claims about machine consciousness, understanding, or phenomenal experience. It is not a contribution to machine learning theory: it proves no new theorems and proposes no new architectures. And while it draws on educational philosophy, it is not primarily a contribution to learning sciences or pedagogy.

Its object is more specific: the epistemic capacities of a particular mathematical structure—what Gibson (1979) would call its *affordances* for knowledge production. The question it asks is not "what are the properties of high-dimensional spaces?" (this is well established) but "what do these properties make possible as conditions for knowledge production?" This question cannot be answered from within any single discipline. Mathematics can characterize concentration of measure (the tendency of distances to become indistinguishable in high dimensions), near-orthogonality, and exponential directional capacity, but it cannot determine what these phenomena mean for epistemology. Philosophy can articulate modes of signification and knowledge production, but without formal engagement with the geometry, such articulations risk remaining metaphorical. Machine learning can demonstrate that generative systems work, but it cannot explain why the medium in which they operate is epistemically productive rather than merely computationally convenient.

The methodology adopted here is therefore deliberately synthetic: it takes rigorously established mathematical properties as its starting point, mobilizes semiotic and constructionist frameworks to interpret their epistemic significance, and develops a unified account—indexical epistemology—that could not emerge from any of these perspectives in isolation. The claim is not interdisciplinary in the weak sense of "drawing on multiple fields"; it is that the object of inquiry itself—the epistemic status of high-dimensional geometry—exists in a space that no single discipline has claimed, and that articulating it requires a form of analysis adequate to its structural character.

## 1.3 From Symbolic Computation to Geometric Navigation: The Paradigmatic Break

The computers and computational tools that surround us—and that have shaped both the practice and the philosophy of computing for over seventy years—operate within a well-defined epistemic regime. Its foundations are three. Turing (1936) established that any effectively computable function can be realized by a finite automaton manipulating symbols on a tape. Shannon (1948) provided the mathematical theory of information as a measurable, transmissible, and encodable quantity, abstracted from meaning. von Neumann (1945) translated these theoretical foundations into a practical architecture: the stored-program computer, in which instructions and data share a common memory and are processed sequentially by a central unit. Together, these contributions constitute what might be called the *symbolic-computational paradigm*: information enters the machine as encoded binary vectors, is processed through rule-governed transformations, and exits as encoded binary vectors. Crucially, semantics remains external to the process. The machine manipulates codes; meaning is assigned by the human interpreter. This is precisely the insight formalized by Searle's (1980) Chinese Room argument and presupposed by the entire tradition of symbolic AI.

What happens at the boundary of a neural network is something for which this paradigm has no adequate description. Consider, following Wolfram (2023), a simple example: the expression "a cat in a party hat" enters a generative system as a sequence of encoded characters—a binary vector of the kind that any Turing machine could process. But the moment this input crosses the threshold of the neural network, it undergoes a transformation that is not merely computational but ontological. Each token is projected into a high-dimensional vector space where its coordinates correspond not to syntactic features—not to letters, bytes, or grammatical categories—but to semantic parameters: dimensions encoding visual appearance, animality, festivity, scale, context, emotional register, and thousands of other features that collectively constitute the concept's position in a space of meanings. The binary input has been translated from a symbolic code into a point in a geometric semantic space.

This translation is the moment at which the high-dimensional space that the present paper analyzes comes into being. It is not a mathematical abstraction imposed by the theorist for purposes of analysis; it arises necessarily from the architecture of neural computation. Every layer of a deep neural network transforms its input into a new high-dimensional representation; every attention mechanism computes geometric relationships—angles, projections, weighted alignments—between vectors in this space. The high-dimensional geometry examined in Section 2 is not a convenient metaphor; it is the operational medium of the system.

This marks a paradigmatic break with the Turing–Shannon–von Neumann tradition—a break that is often obscured by a contingent fact of engineering practice. Today's neural networks are, of course, implemented on von Neumann architectures: their matrix multiplications are executed by sequential processors, their weights stored in addressable memory, their training orchestrated by conventional algorithms. This implementation dependency, however, is a historical contingency, not a conceptual necessity. A neural network does not require a von Neumann machine; it can, in principle, operate on entirely different physical substrates—optical processors, neuromorphic chips, analog circuits, or substrates not yet invented. The von Neumann machine emulates the neural process; it does not constitute it. The relationship is analogous to that between a weather simulation and the weather itself: the simulation runs on a Turing-complete machine, but the atmospheric dynamics it models are not Turing computations.

This distinction has consequences that extend beyond engineering into epistemology. A researcher who understands deep learning technically—who can derive backpropagation, implement transformer architectures, and optimize training regimes—but whose computational thinking (Wing, 2006) remains rooted in the Turing–von Neumann paradigm faces a subtle but consequential conceptual limitation. Such a thinker may see the neural network as an elaborate algorithm executing on a conventional machine and conclude that nothing fundamentally new is happening—that the system is, at bottom, "just" manipulating symbols faster and at greater scale. This conclusion mistakes the medium of implementation for the mode of operation. What is new is not the hardware but the epistemic regime: the transition from rule-governed symbol manipulation to geometric navigation through learned semantic spaces.

It is this transition that the present paper sets out to analyze. The paper proceeds as follows. Section 2 examines four structural properties of high-dimensional geometry—concentration of measure, near-orthogonality, exponential directional capacity, and manifold regularity—and their epistemic implications. Section 3 analyzes the migration of meaning from metric distance to positional orientation. Section 4 articulates the central claim of the paper: that generative AI constitutes a third mode of knowledge production, distinct from both symbolic reasoning and statistical recombination, and provides a precise account of what knowledge production means in this geometric regime. Section 5 develops the epistemology of this third mode, drawing on Peirce's semiotics and Papert's constructionism. Section 6 addresses objections and clarifications. Section 7 examines the question of structural agency. Section 8 concludes with reflections on the geometrization of epistemic structure.

## 2   High-Dimensional Geometry: Structural Properties and Epistemic Consequences

Generative models operate in vector spaces with dimensions ranging from hundreds to hundreds of thousands. These spaces are not merely large versions of familiar Euclidean geometry; they exhibit qualitatively different structural properties that fundamentally alter what it means for something to be "near," "similar," or "related." As Vershynin (2018) demonstrates rigorously, high-dimensional probability reveals phenomena that are not extrapolations of low-dimensional intuition but genuine structural novelties emerging only when dimensionality crosses certain thresholds.

The epistemological significance of these properties has not yet been systematically examined. Computer scientists treat them as computational resources; mathematicians study them as abstract phenomena; machine learning practitioners exploit them as engineering affordances. What remains to be developed is an account of their epistemic implications—of what they make possible as conditions for knowledge production.

Four structural properties are decisive: concentration of measure (Milman, 1971; Ledoux, 2001), which eliminates metric contrast; near-orthogonality (Vershynin, 2018), which establishes representational independence as a default; exponential directional capacity (Kabatyansky & Levenshtein, 1978; Milman & Schechtman, 1986), which provides combinatorial richness far exceeding any training set; and manifold regularity (Bengio et al., 2013; Tenenbaum et al., 2000), which channels this richness into coherent navigable structure. Each is examined in turn below.

### 2.1 Concentration of Measure: The Collapse of Metric Intuition

Concentration of measure describes a phenomenon that is central to understanding why high-dimensional spaces behave so differently from the spaces of everyday experience. First identified by Milman (1971) in the context of asymptotic geometry of Banach spaces—extending earlier observations by Lévy—and later developed into a comprehensive theory (Ledoux, 2001), this phenomenon addresses a simple core question: if we place a collection of random points in a space, how are they distributed relative to the origin? In two or three dimensions, random points scatter widely—some land close to the origin, others far away, and the spread is substantial. In high-dimensional spaces, something radically different occurs: random points cluster in a thin shell at a nearly fixed distance from the origin. The variation in their distances, which is large in low dimensions, shrinks to near-zero as dimensionality grows.

This phenomenon can be stated precisely (see Vershynin, 2018, Chapter 3). Consider a random vector **x** in an $n$-dimensional space, where each of its $n$ components is drawn independently from a standard normal distribution. The length of this vector measures its distance from the origin. The key result is surprisingly simple: in high dimensions, this length is almost exactly predictable. It equals approximately $\sqrt{n}$ (the square root of the number of dimensions), and deviations from this value become exponentially unlikely as $n$ grows. More formally:

$$P\big(\big|\|\mathbf{x}\| - \sqrt{n}\big| > \varepsilon\sqrt{n}\big) \leq 2\,e^{-cn\varepsilon^2}.$$

The formula says exactly what the preceding sentence says, but with mathematical precision: the probability ($P$) that the vector's length ($\|\mathbf{x}\|$) deviates from $\sqrt{n}$ by more than a fraction $\varepsilon$ shrinks exponentially with the dimension $n$. The constant $c$ is a fixed positive number. A reader unfamiliar with the notation can safely retain just the plain-language version: in a 768-dimensional space, virtually every random vector has almost exactly the same length. The shell in which these vectors concentrate becomes, proportionally, thinner and thinner as dimensionality increases—not gradually, but exponentially fast. As illustrated in Figure 1, the distribution of vector norms shifts from a widespread in low dimensions to a sharp peak around √n in high dimensions.

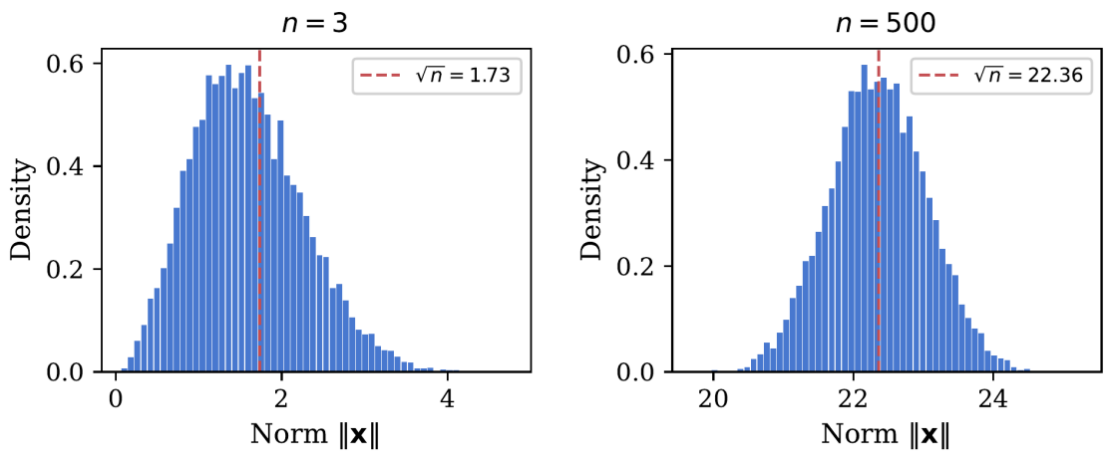

**Figure 1:** Concentration of measure illustrated. Left: In n=3 dimensions, vector norms are widely dispersed. Right: In n=500 dimensions, virtually all norms cluster tightly around √n ≈ 22.4. The dashed line marks √n.

This is not a statistical curiosity but a structural transformation of the space itself. In low dimensions, the distance between two random points carries rich information: proximity indicates similarity, remoteness indicates dissimilarity. This intuitive metric semantics underwrites classical notions of clustering, classification, and analogy. In high dimensions, this semantics collapses. Aggarwal et al. (2001) showed that for any given point in a high-dimensional dataset, the ratio of the distance to its farthest neighbor to the distance to its nearest neighbor converges to 1 as dimensionality increases—meaning that, from the perspective of any single point, all other points become approximately equidistant.

This result is deeply counter-intuitive. In everyday experience, some things are close and others are far away—and this difference is informative. We rely on it constantly: the nearest neighbor is more relevant than a distant one, a cluster of nearby points indicates a category, and outliers are recognizable precisely because they are remote. Aggarwal et al. demonstrated that in high-dimensional spaces, this entire logic dissolves: the farthest point in a dataset is barely farther than the nearest one. When all distances become nearly equal, distance itself ceases to be a useful carrier of meaning.

**Epistemic implication.** The collapse of metric contrast is not merely a technical inconvenience requiring better distance measures. It is an epistemological event: the mode of knowledge that depends on distance—classification by proximity, identity by metric similarity—ceases to function. High-dimensional space becomes what we might call *statistically rigid*: globally uniform in its metric properties, yet locally structured in ways that require entirely different modes of access. This rigidity is the first indication that high-dimensional geometry demands a new epistemological framework—one in which meaning cannot be grounded in magnitude alone.

Floridi (2011) argues that the level of abstraction at which a system is analyzed determines what counts as informative within that system. A change in the level of abstraction is not merely a methodological choice; it can reveal structures invisible at other levels—or render previously informative features meaningless. Concentration of measure is precisely such a shift. It does not merely complicate distance-based reasoning; it invalidates it as an epistemic strategy. When all points in a space are approximately equidistant, the question "how far is $A$ from $B$?" no longer yields useful distinctions. The informative question becomes a different one: "in what direction does $A$ lie relative to $B$?" This is a forced migration of epistemic strategy—from metric to directional semantics, from magnitude to orientation—imposed not by theoretical preference but by the geometry of the space itself.

## 2.2 Near-Orthogonality: Independence as Structural Default

The second structural property concerns the relationship between pairs of vectors. If concentration of measure tells us that individual vectors all have approximately the same length, near-orthogonality tells us something equally striking about how vectors relate to one another: in high-dimensional spaces, any two randomly chosen vectors are almost certainly perpendicular.

The mathematical result is straightforward and well established (see, e.g., Vershynin, 2018, Chapter 3). Take two vectors chosen at random from the surface of a high-dimensional sphere. The dot product of two vectors—a single number that measures how much they point in the same direction—ranges from $-1$ (exactly opposite) through $0$ (perfectly perpendicular) to $+1$ (exactly aligned). For random vectors in high-dimensional spaces, this dot product is on average exactly zero (meaning: no alignment at all), and the spread around zero equals $1/n$, where $n$ is the number of dimensions. In notation:

$$\mathbb{E}[\mathbf{x} \cdot \mathbf{y}] = 0, \qquad \text{Var}(\mathbf{x} \cdot \mathbf{y}) = \frac{1}{n}.$$

The formula confirms the verbal statement: the expected alignment ($\mathbb{E}$) is zero, and the variance (Var)—the typical size of random fluctuations around zero—is $1/n$. In 768 dimensions, this variance is approximately 0.0013: the typical alignment between any two random vectors is vanishingly small. They are, for all practical purposes, perpendicular. Figure 2 shows how the distribution of cosine similarity collapses around zero as dimensionality grows.

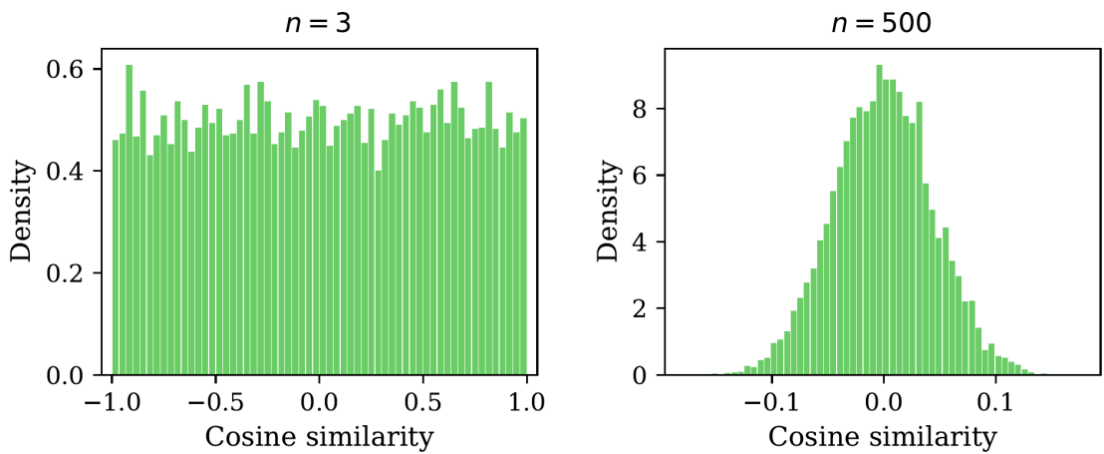

**Figure 2:** Near-orthogonality illustrated. Left: In n=3 dimensions, cosine similarity between random vector pairs is uniformly distributed across [−1,+1]. Right: In n=500 dimensions, the distribution collapses sharply around zero, confirming that random pairs are nearly orthogonal by default.

Why does the angle between vectors matter? In embedding spaces, the angle captures something fundamental: the degree to which two representations share structure. When two vectors point in the same direction—when the angle between them is small—they respond to similar contexts, activate in similar circumstances, and carry related meanings. When they are orthogonal—when the angle between them is 90 degrees—they share nothing; each varies entirely independently of the other. The cosine of this angle, ranging from 1 (perfect alignment) through 0 (complete independence) to −1 (opposition), thus serves as a natural measure of semantic relatedness. This is why near-orthogonality has such profound consequences: it determines the baseline condition of how concepts relate to one another in the space.

As the dimension increases, almost all pairs of vectors become nearly orthogonal—that is, almost all pairs of representations become semantically independent by default. In low-dimensional spaces, orthogonality is a special relationship that must be engineered or discovered. In high-dimensional spaces, it is the default condition. Semantic relatedness—the situation in which two vectors share a significant degree of alignment—becomes the exception that requires explanation.

**Epistemic implication.** Near-orthogonality provides the geometric foundation for what might be called *representational autonomy*—the capacity of a space to host an enormous number of semantically independent directions without mutual interference. In the language of information theory, high-dimensional spaces possess extraordinarily high channel capacity: they can encode vast numbers of independent signals simultaneously. But the epistemological point runs deeper than information-theoretic efficiency. In symbolic systems, independence between representations must be *declared*: variables are distinct because they are syntactically different tokens. In high-dimensional embedding spaces, independence is *geometric*: representations are autonomous because the space itself provides sufficient room for them to coexist without interference. This is not engineered independence—it is structural. The space does not need to be told which concepts are distinct; its dimensionality makes distinctness the default condition.

What must be learned is not separation but connection. Since high-dimensional space makes independence the default—since any two randomly chosen representations will almost certainly be orthogonal—the space does not need to learn how to keep concepts apart. Distinctness comes for free. What is rare, and therefore informative, is the opposite: when two vectors do share significant alignment, this is not accidental but reflects genuine structure in the data. The training process, in effect, carves out meaningful relationships against a background of universal independence. It is these learned alignments—departures from the orthogonal default—that carry semantic content.

## 2.3 Exponential Directional Capacity: Combinatorial Richness

The third property addresses a question that is fundamental to understanding why generative AI can be so expressive: how many independent directions can a space accommodate? The answer depends dramatically on dimensionality. One way to think about this is to ask: how many vectors can we pack into a space such that no two of them are significantly aligned—that is, each pair is approximately orthogonal? In geometric terms, this is equivalent to asking how many non-overlapping "caps" of a given size can fit on the surface of a high-dimensional sphere—a problem known in mathematics as sphere packing.

The answer, established through classical results in geometric combinatorics (Kabatyansky & Levenshtein, 1978; Milman & Schechtman, 1986), can be stated simply: the number of approximately independent directions does not grow linearly with dimension (adding one direction per new dimension) but exponentially. In notation:

$$N(\varepsilon) \sim e^{cn}.$$

Here $N$ is the number of nearly independent directions, $n$ is the number of dimensions, and $c$ is a positive constant that depends on how strictly we define "independent" (the parameter $\varepsilon$). A reader unfamiliar with exponential notation can substitute a concrete image: in three dimensions there are exactly three independent directions ($x$, $y$, $z$); in 768 dimensions the number of nearly independent directions exceeds the number of atoms in the observable universe by many orders of magnitude. Each such direction can, in principle, encode a separate semantic distinction. As schematically illustrated in Figure 3, this capacity scales exponentially rather than linearly with dimension.

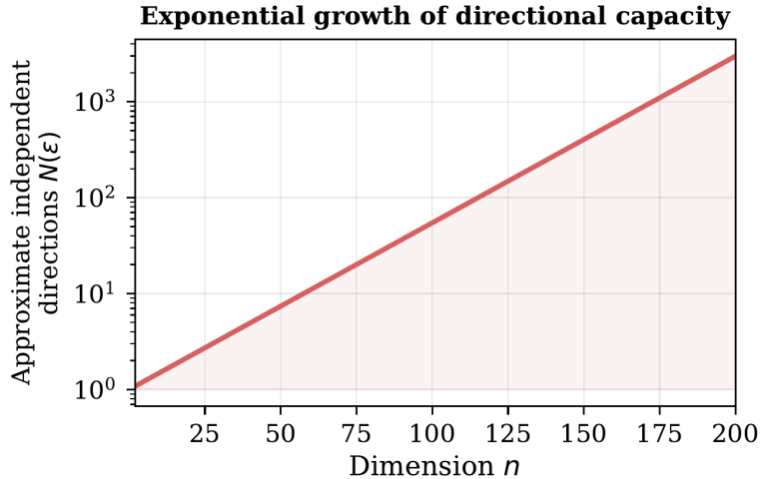

**Figure 3:** Exponential growth of directional capacity. The number of approximately independent directions N(ε) grows exponentially with dimension n (log scale on y-axis). Even modest increases in dimensionality produce enormous increases in the space of available semantic distinctions.

**Epistemic implication.** Exponential directional capacity means that the space of geometrically admissible configurations vastly exceeds the cardinality of any training dataset. This has a profound consequence for the question of novelty.

This point warrants emphasis, as it addresses one of the most widespread assumptions in the current debate. Critics who argue that generative systems can only "recombine" training data implicitly assume a discrete, combinatorial model of possibility—as if the system's repertoire were a finite deck of cards that can only be reshuffled. But high-dimensional geometry does not deal in cards. In a continuous space with exponential directional capacity, the set of geometrically accessible configurations is not merely larger than the training set; it is of a fundamentally different cardinality—as different as the real number line is from the integers. The training data do not define the boundary of what the system can produce; they define the manifold along which production becomes possible.

Novelty in such a space is not exceptional, not accidental, and not illusory. It is structurally inevitable: a mathematical consequence of the geometry itself.

The question is not whether generative systems produce novel outputs, but whether the specific form of novelty they produce is epistemically meaningful—and this depends on the fourth structural property.

## 2.4 Manifold Regularity: Local Coherence within Global Richness

The three properties examined so far—concentration, near-orthogonality, and exponential capacity—describe the ambient high-dimensional space in its raw form: vast, statistically rigid, and combinatorially rich. But generative models do not operate in this raw space uniformly. Their training data, and the meaningful structures they learn, do not fill the entire volume of the space; they occupy only a thin, structured region within it. This observation is formalized in what has become known as the *manifold hypothesis* .

The concept of a manifold can be understood through a familiar analogy. The surface of the Earth is a two-dimensional manifold embedded in three-dimensional space: locally, it looks flat—flat enough to build houses and draw maps—but globally, it curves into a sphere. The key insight is that while the Earth's surface exists in three dimensions, life on it is effectively two-dimensional: we navigate using only latitude and longitude, not altitude. A manifold, in general, is a structure that behaves the same way—locally simple and smooth, globally complex and curved, and always of lower dimensionality than the space that contains it.

The manifold hypothesis (Bengio et al., 2013; Tenenbaum et al., 2000; Roweis & Saul, 2000) holds that meaningful data—natural images, coherent text, plausible speech signals—occupy such structures within high-dimensional embedding spaces. A 768-dimensional space can host an astronomical number of possible vectors, but natural language does not use them all: the vectors that correspond to meaningful utterances lie on a manifold of far fewer effective dimensions, winding through the ambient space like a complex surface through a higher-dimensional void.

These manifolds possess geometric regularity. They are locally smooth—nearby points on the manifold represent similar meanings. They have consistent directional structure—one can trace continuous paths along them without abrupt jumps. And they curve in measurable ways—the relationships between meanings change gradually, not discontinuously. The training process in generative models can be understood as the progressive discovery and parameterization of these manifolds—learning not where all points are, but where the meaningful ones lie. Figure 4 illustrates this structure: training instances lie on a curved manifold, and generative trajectories move through regions not explicitly sampled.

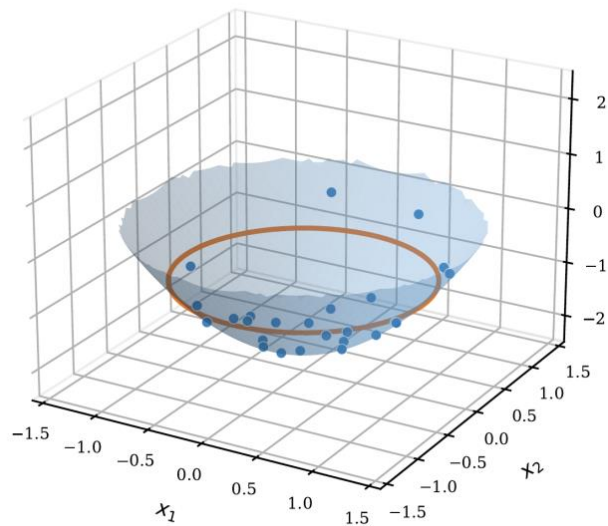

**Figure 4:** Conceptual manifold traversal. Training instances (blue dots) lie on a smooth, lower-dimensional manifold (blue surface) embedded in higher-dimensional ambient space. The orange contour marks a region of the manifold accessible through continuous navigation.

**Epistemic implication.** The first three properties, taken alone, might seem to paint a troubling picture. Concentration of measure tells us that distance is uninformative. Exponential capacity tells us that the space contains unimaginably many possible configurations. Together, they suggest a space that is simultaneously rigid and vast—and vastness without structure is simply chaos. If a generative model could wander freely through all of high-dimensional space, its outputs would be noise, not language.

Manifold regularity resolves this tension. The manifold acts as a geometric constraint that channels the combinatorial richness of the space into coherent, navigable trajectories. A generative model does not wander through the full volume of 768-dimensional space; it moves along the manifold—the thin, structured surface where meaningful configurations reside. Because this surface is smooth, nearby points represent similar meanings. Because it is continuous, the model can trace paths along it without abrupt jumps. And because it is lower-dimensional than the ambient space, the effective landscape is manageable even though the surrounding space is vast. The manifold is what turns raw geometric possibility into structured generativity.

In philosophical terms, the manifold functions as what Simondon (1958/2020) called a *transductive milieu*—a medium in which new structure emerges not by assembling pre-given parts but by progressively resolving tensions and potentials inherent in the medium itself. Each point on the manifold is determined not by its absolute coordinates in ambient space but by its differential relations to its neighbors—by the local geometry of the surface at that point. A generative model traversing this manifold does not retrieve stored answers; it individuates new configurations through its interaction with the manifold's structure. This is precisely the kind of medium that makes navigational knowledge—knowledge constituted through movement rather than storage—possible.

**A note on mathematical assumptions.** The results presented in this section are asymptotic: they describe the behavior of high-dimensional spaces in the limit as dimensionality grows. They rely on idealizations—Gaussian distributions, statistical independence of components—that real embedding spaces do not satisfy exactly. Practical embedding dimensions (768 for BERT-base, 1536 for GPT-3, 12288 for GPT-4) are finite, and the vectors they contain are shaped by training dynamics that introduce complex dependencies. Nevertheless, empirical research consistently confirms that the structural tendencies described here—concentration of norms, near-orthogonality of random pairs, and the manifold organization of meaningful data—are robustly observed in practice at these dimensionalities (Bengio et al., 2013; Vershynin, 2018). This paper treats these properties not as exact theorems about specific models but as structural tendencies of high-dimensional geometry that shape the epistemic landscape in which generative systems operate.

# 3 From Geometry to Positional Knowledge: The Migration of Structure

The four properties outlined above—concentration, near-orthogonality, exponential capacity, and manifold regularity—are not isolated mathematical curiosities. Together, they constitute a structural transformation in the conditions of knowledge production. To see this transformation clearly, it is helpful to trace what happens to the concept of similarity as we move from low to high dimensions.

The connection between the collapse of distance and the primacy of orientation is not merely conceptual; it is mathematically precise. For two vectors **x** and **y**, the squared distance between them decomposes according to the law of cosines:

$$\| \mathbf{x} - \mathbf{y} \|^2 = \| \mathbf{x} \|^2 + \| \mathbf{y} \|^2 - 2 \| \mathbf{x} \| \| \mathbf{y} \| \cos\theta,$$

where $\theta$ is the angle between them. In high-dimensional spaces, concentration of measure forces both norms to converge on the same value (approximately $\sqrt{n}$), while near-orthogonality forces $\cos\theta$ toward zero. The result is that $\| \mathbf{x} - \mathbf{y} \|^2 \approx 2n$ for virtually any pair of random vectors—all distances become indistinguishable. This is exactly the Aggarwal et al. result, now derived rather than merely observed. But the decomposition reveals something further: while absolute distance has lost its discriminative power, the angular component—$\cos\theta$—retains it. Against a background of near-universal orthogonality, even a small deviation from zero cosine becomes a strong signal. Cosine similarity works not despite the curse of dimensionality but because of it: the geometric default of orthogonality provides a silent background against which genuine semantic alignment becomes detectable—like a whisper in a perfectly quiet room.

In low-dimensional geometry, similarity is grounded in distance. Two points are "similar" if they are metrically close; they are "different" if they are far apart. This metric semantics underwrites an entire epistemological tradition: classification, clustering, nearest-neighbor reasoning, and even the philosophical notion of family resemblance (Wittgenstein, 1953) all presuppose that similarity is a function of proximity in some appropriate space.

In high-dimensional regimes, this metric semantics breaks down. As concentration of measure eliminates global distance contrast, orientation becomes more informative than magnitude. Cosine similarity—which measures the angle between vectors rather than the distance between them—replaces Euclidean distance as the operative measure of semantic relatedness in virtually all modern embedding systems (Mikolov et al., 2013; Radford et al., 2021). This is not merely a technical substitution of one metric for another; it is a shift in the mode of signification. Meaning migrates from *what* a vector is (its magnitude, its absolute position) to *where* a vector points (its orientation, its relational alignment with other vectors).

An embedding vector, in this light, is not a symbolic definition. It does not encode a concept through a list of necessary and sufficient conditions, nor through a set of logical predicates. It is a location within a structured relational field. Its interpretive force arises entirely from its position relative to other vectors—from the angles it subtends, the neighborhoods it inhabits, the manifold regions it activates. As Wittgenstein (1953) observed in a different context: "The meaning of a word is its use in the language." In high-dimensional embedding spaces, the meaning of a vector is its position in the geometry.

This positional semantics represents what Floridi (2011) would call a shift in the level of abstraction—not a loss of structure but a migration of structure from one register to another. Knowledge becomes navigational rather than representational: to "know" something in an embedding space is not to possess a symbolic description of it but to be able to locate it, to orient toward it, to traverse the manifold paths that connect it to related regions. Structure has not disappeared. It has migrated—from the explicit, human-readable architecture of logical formulae to the implicit, high-dimensional architecture of geometric relations.

# 4 The Third Mode of Knowledge Production: Geometric Novelty Beyond Reasoning and Recombination

The central claim of this paper is that generative AI constitutes a *third mode of knowledge production*—fundamentally distinct from the two modes that have dominated both computational practice and philosophical reflection. The first is *symbolic reasoning*: rule-governed manipulation of discrete representations, the epistemic regime of classical AI, formal logic, and the Turing–von Neumann tradition. The second is *statistical recombination*: pattern extraction from data through correlation, the regime of classical machine learning, Bayesian inference, and connectionist pattern-matching. The third mode—which this section sets out to characterize—is *geometric navigation*: the production of coherent, contextually appropriate novelty through structured traversal of learned manifolds in high-dimensional space.

This claim rests on the geometric foundations established in Sections 2 and 3. The four structural properties examined there are not independent curiosities; they work together to create a specific structural situation. Concentration of measure eliminates naive distance-based reasoning. Near-orthogonality provides an enormous reservoir of independent directions. Exponential capacity ensures that the number of available configurations dwarfs any finite dataset. And manifold regularity ensures that this vastness is not chaotic but geometrically organized—navigable, coherent, and locally smooth. The question is: what do these properties, taken together, make possible?

Consider what happens when a generative model produces an output. It does not select an item from a stored list, nor does it assemble one from memorized fragments. It traces a trajectory through the learned manifold—a continuous path shaped by the input context, the model's parameters, and the geometric constraints of the space. The endpoint of this trajectory is a point on the manifold: a geometrically admissible configuration—admissible because it is consistent with the manifold's local structure, its smoothness, its curvature. Such a point need not coincide with any training example. It need only lie on or near the manifold that the training process has discovered.

The central structural claim of this paper can now be stated precisely: in sufficiently high-dimensional spaces exhibiting concentration, near-orthogonality, exponential directional capacity, and manifold regularity, the set of geometrically admissible configurations—points reachable through continuous navigation of the learned manifold—vastly exceeds the set of training instances from which that manifold was learned. The manifold contains more than what was used to build it.

This is not a merely quantitative observation ("there are more possible outputs than training inputs"). It is a structural claim about the nature of the excess. The geometrically admissible configurations are not random points scattered through ambient space; they are positions on the learned manifold that are consistent with its local structure—its tangent directions, its curvature, its neighborhood relations. Generative systems produce novelty not by randomly sampling the ambient space (which would yield noise) nor by memorizing and replaying training instances (which would yield copies), but by traversing the manifold through regions that are geometrically coherent yet not explicitly represented in the training data.

This account addresses what is perhaps the most consequential unexamined dichotomy in the current debate about generative AI. The dichotomy, rarely stated explicitly but pervasive in its influence, assumes that generative outputs must fall into one of two categories: either they are products of genuine reasoning—in which case the system must possess something like understanding, intentionality, or cognitive depth—or they are products of statistical recombination—in which case novelty is illusory and the system is, at best, a sophisticated parrot. This binary has shaped public discourse, philosophical argument, and policy discussion alike. It forces a choice: attribute too much to the machine, or dismiss it entirely.

High-dimensional geometry reveals that this choice is unnecessary. There exists a third possibility that neither horn of the dilemma acknowledges: structurally enabled novelty grounded in geometric organization. The system neither "understands" in the symbolic-inferential sense nor merely "copies"; it navigates a geometrically structured space in which coherent novelty is a natural consequence of continuous traversal. This is not a compromise position or a diplomatic middle ground. It is a fundamentally different account of what generative systems do—one that becomes visible only when we attend to the medium in which they operate.

The implications for the debate around AI creativity are significant. Boden (2004) distinguished three forms of creativity: combinational (novel combinations of familiar ideas), exploratory (traversal within an accepted conceptual space), and transformational (alteration of the space itself). Generative AI, on this account, operates primarily in the exploratory mode—but with a crucial difference. The "conceptual space" being explored is not a set of discrete rules or conventions but a continuous manifold in a high-dimensional vector space. The richness of this space—its exponential directional capacity—means that exploratory creativity alone can produce outputs of striking novelty, without requiring the transformational step of altering the space's structure.

The geometry itself is the source of creative possibility. This is what becomes visible once we move beyond the dichotomy between reasoning and recombination.

The foregoing analysis substantiates the claim advanced at the opening of this section. Geometric navigation—the third mode of knowledge production—is a mode in which structure resides not in explicit rules nor in memorized patterns but in the geometry of a high-dimensional space; in which meaning is positional rather than definitional; and in which novelty arises through traversal rather than through inference or copying. This mode has no established name in the current discourse, precisely because the prevailing binary framework does not readily accommodate it. The indexical epistemology developed in the following section provides the conceptual vocabulary it requires.

But calling this a mode of knowledge production rather than a sophisticated form of interpolation requires justification. The objection is predictable: "You have redescribed statistical interpolation in philosophical language. The model simply fills in gaps between training points on a smooth surface—a mathematically trivial operation, not an epistemic one." This objection must be met directly.

Four considerations show why the interpolation framing is inadequate. First, classical interpolation operates within a known space—between data points whose coordinates are given in a pre-existing coordinate system. In generative AI, the manifold is not given; it is discovered through training. The model does not fill gaps on a known map; it constructs the map itself. This is closer to cartographic exploration than to connecting dots—and the cartographer who maps an uncharted continent produces knowledge, even though the continent existed before the map.

Second, the scale of what lies between training points is qualitatively different in high-dimensional spaces. In low-dimensional interpolation, the gaps between data points are narrow, and interpolated values are predictable from local neighborhoods. In spaces with exponential directional capacity, the regions between training instances are geometrically vast—containing configurations that are consistent with the manifold's structure but not derivable from any local subset of training data. This is not gap-filling; it is traversal through territory that is wider than the data from which it was discovered.

Third, the normative criterion for meaningful novelty is provided by the manifold itself. Not every geometrically admissible configuration is meaningful, and the manifold is not arbitrary: it is a trace of the semantic structure of human linguistic and perceptual practice, shaped by millions of contextual regularities. A configuration is meaningful not because a human approves it post hoc, but because it satisfies geometric constraints that were learned from structured human practice. The manifold encodes normativity geometrically.

Fourth, and most fundamentally: what kind of knowledge is this? It is not propositional knowledge ("knowing that"), nor procedural knowledge in the classical sense ("knowing how" through explicit rules). This paper proposes the term *navigational knowledge*: the capacity to produce contextually coherent, geometrically admissible configurations through structured traversal of a learned manifold. Navigational knowledge is characterized by three constitutive features: it is *positional* (determined by location in a structured space, not by definition); it is *enactive* (produced through traversal, not retrieved from storage); and it is *bounded* (constrained by manifold structure, not arbitrary). The critical distinction is this: interpolation is an operation within a known space; navigational knowledge is a capacity that arises from the interaction between an agent and the structure of the space it inhabits. Interpolation does not produce an agent; navigation does.

Establishing the existence of a third mode, however, is not the same as understanding it. Two questions remain that geometric analysis alone cannot answer. The first is semiotic: how does positional meaning work? To say that a vector "means" something by virtue of its position is, without further analysis, a metaphor rather than a theory. What is needed is a precise account of the mode of signification at work in embedding spaces—an account that can distinguish this mode from the symbolic signification of classical AI and explain why positional relation generates meaning at all. The second question is epistemological in a broader sense: what does this shift imply for how knowledge is constructed, transmitted, and learned? A new mode of knowledge production demands a re-examination of the frameworks—educational, institutional, philosophical—through which we understand knowledge itself. Section 5 addresses both questions: drawing first on Peirce's semiotics to provide the missing theory of signification, and then on Papert's constructionism to articulate the educational and epistemological consequences.

# 5   The Epistemology of the Third Mode of Knowledge Production

## 5.1 Indexical Signification: How Position Generates Meaning

The epistemological implications of the structural analysis developed in Sections 2–4 can be clarified through a semiotic lens. In Peirce's triadic classification of signs (CP 2.243–2.253), three modes of signification are distinguished:

*Symbols* signify through convention and rule. The word "tree" refers to trees by virtue of a linguistic convention, not by virtue of any intrinsic connection between the word and its referent. Symbolic signification is the dominant mode of classical AI: LISP atoms, logical predicates, and ontological categories all operate through conventionally established correspondences.

*Icons* signify through resemblance. A photograph of a tree signifies a tree by sharing visual properties with it. Iconic signification plays a role in some forms of representation learning, but it is not the primary mode of operation in language models or embedding systems.

*Indices* signify through existential connection—through a real, non-conventional relation between sign and object. A weathervane points to the wind's direction not by convention or resemblance but by being physically connected to the wind. Smoke indicates fire not because we have agreed that it should, but because fire causally produces smoke. The index, as Peirce (CP 2.248) emphasizes, "is a sign which refers to the Object that it denotes by virtue of being really affected by that Object."

The central semiotic claim of this paper is that generative AI operates *indexically*. This claim extends an important insight first articulated by Weatherby & Justie (2022), who argued that neural networks constitute a form of "indexical AI" that "points rather than describes," positioning their semiotic function between symbolic heuristics and genuine intelligence. The present analysis provides the geometric foundation that explains *why* this indexicality is epistemically productive: it is the structural properties of high-dimensional spaces—concentration, near-orthogonality, exponential capacity, and manifold regularity—that make indexical signification not merely possible but the dominant mode of meaning-generation in these systems.

Embedding vectors acquire their meaning not through symbolic convention (they are not assigned meanings by a programmer) nor through iconic resemblance (they do not "look like" what they represent). They signify through their positional relation to the structured field in which they are embedded. An embedding vector for the concept "king" does not contain a symbolic definition of kingship; it occupies a position in vector space that stands in specific geometric relations to the positions occupied by "queen," "man," "woman," "ruler," and thousands of other concepts. Its meaning is constituted by these relations—by its existential placement within a relational field.

The nature of this indexicality can be captured through a simple formulation: an embedding vector means what it means *here* (by virtue of its position in a structured space), *now* (in the specific context established by the current prompt and attention configuration), and *with these neighbors* (through its geometric relations to surrounding vectors). These three aspects—positional, contextual, and relational—are precisely what distinguish indexical from symbolic signification. A symbolic definition of "king" remains the same regardless of where it is stored, when it is accessed, or what surrounds it. An embedding vector for "king" is constituted by its here-now-with: its meaning shifts as the context changes, as attention reconfigures the relational landscape, as different regions of the manifold are activated. This is indexicality in its strongest Peircean sense: meaning that arises from the sign's actual situation within a structured field, not from a convention imposed upon it.

This indexicality is not metaphorical. Peirce's definition of the index requires a "real connection" between sign and object—a relation that is not arbitrary but grounded in the actual structure of the world. In embedding spaces, the connection is grounded in the statistical structure of the training corpus, which is itself a trace of the structure of human linguistic practice. The position of "king" in vector space is not arbitrary; it is the result of millions of contextual co-occurrences that constrain the vector's placement through gradient descent. The embedding is, in a precise Peircean sense, an index: a sign whose position is determined by a real causal-statistical process, not by arbitrary convention.

An *Indexical Epistemology of High-Dimensional Spaces*, therefore, describes a mode of knowledge production characterized by three features:

*Positional meaning.* Meaning is relational and positional rather than definitional. A concept is "known" not by possessing a symbolic description of it but by occupying the right position in a structured geometric field.

*Geometric coherence.* Coherence arises from the constraints imposed by manifold structure, concentration of measure, and near-orthogonality—not from logical consistency or syntactic well-formedness.

*Navigational novelty.* Novelty emerges through structured navigation along the manifold rather than through combinatorial manipulation of discrete symbols.

Knowledge, in this framework, is not retrieved from storage; it is enacted through navigation. This formulation resonates with Varela et al.'s (1991) enactive epistemology, in which cognition is understood as embodied action rather than abstract computation. But where enactivism grounds knowledge in biological embodiment, indexical epistemology grounds it in geometric embodiment—in the specific structural affordances of high-dimensional spaces.

## 5.2 Constructionist Implications: From Symbolic Microworlds to Geometric Navigation

The indexical epistemology proposed above finds a productive counterpart in Papert's constructionist tradition. Papert (1980) argued that knowledge is not passively absorbed but actively constructed through engagement with microworlds—computationally mediated environments in which learners build, test, and revise their understanding through direct manipulation. His fundamental insight was epistemological, not merely pedagogical: the medium of knowledge construction shapes the knowledge that can be constructed within it. But Papert's microworlds were symbolic environments—mediated by discrete commands, governed by explicit rules, operating in visualizable low-dimensional spaces. The Logo turtle constructed mathematical knowledge by following symbolic instructions through a two-dimensional plane. Generative AI operates in a fundamentally different regime: its "microworld" is a high-dimensional vector space, non-visualizable, governed by geometric constraints rather than explicit rules, navigated through continuous activation rather than discrete command. The mode of construction has shifted from programming to navigation, from explicit rule-following to positional sensitivity.

Levin et al. (2025a,b) trace this shift through the broader trajectory of constructionism's evolution across three digital epochs, arguing that generative AI systems transform the constructionist tool from an *object to think with* into an *agent to think with*—an interactive partner that does not think symbolically but navigates high-dimensional manifolds. The fundamental constructionist insight—that knowledge is actively built through engagement with structured environments—survives this transition. But the semiotic register of enactment shifts: from symbolic in Papert's case to indexical in the case of generative AI.

The present analysis extends this trajectory by specifying what kind of medium high-dimensional space is: one in which knowledge is positional rather than definitional, coherence is geometric rather than logical, and novelty arises through navigation rather than through combinatorial assembly. Constructionism, in this light, does not need to be abandoned in the age of generative AI; it needs to be geometrized.

## 6   Objections and Clarifications

The framework developed in this paper—a third mode of knowledge production, grounded in high-dimensional geometry and articulated through indexical epistemology— departs significantly from established positions in both philosophy of AI and machine learning theory. Several objections can be anticipated. This section addresses the four most fundamental.

**Objection 1: "Isn't this just interpolation?"** This objection was addressed in detail in Section 4, but its core can be restated concisely. Interpolation is a mathematical operation that fills gaps between known points in a known space. What generative models do is structurally different in three respects: the space itself (the manifold) is not given but discovered; the "gaps" between training points in high-dimensional spaces are not narrow bands but geometrically vast regions with their own structure; and the traversal is context-sensitive and non-deterministic, producing different outputs for different inputs even when navigating the same region of the manifold. The interpolation framing treats the manifold as a lookup table with smooth edges. The navigational framing treats it as a structured environment through which an agent moves. The difference is not terminological—it is the difference between a mathematical operation and an epistemic capacity.

**Objection 2: "Doesn't the manifold presuppose the very semantic structure you attribute to geometry?"** This is a subtle and important objection. It holds that the manifold's structure is inherited from the training data, which is itself semantically organized; therefore, the semantics reside in the data, not in the geometry. The objection is partly correct: the manifold is indeed shaped by training data, and that data reflects the structure of human linguistic and perceptual practice. But the objection does not address the crucial further point: the manifold's geometric properties—its smoothness, curvature, dimensionality, and connectivity—generate possibilities that are not present in the data themselves. The training data are discrete points; the manifold is a continuous structure that exceeds them. An analogy clarifies: the bed of a river is shaped by the water that flows through it, but the form of the riverbed determines trajectories that the original water never followed. To say that the semantics are "in the data" is like saying the river's future course is "in" the water that carved its bed. It is true causally but inadequate structurally: the geometry, once formed, has its own productive logic.

**Objection 3: "Why call it epistemology rather than representational geometry?"** This objection accepts the geometric analysis but questions the philosophical framing: why not simply call this "the geometry of learned representations" and leave epistemology out of it? The answer lies in the question being asked. A purely geometric account describes *what* the properties of the space are: its dimensionality, its curvature, its statistical regularities. An epistemological account asks what these properties *make possible* as conditions for knowledge production. The difference is between studying an object and studying what the object enables. Mathematics can tell us that near-orthogonality holds in high dimensions; it cannot tell us that this property transforms the conditions under which meaning is generated. Floridi's philosophy of information, Peirce's semiotics, and Papert's constructionism are not decorative additions to a geometric description; they are the analytical tools required to articulate consequences that geometry alone cannot name. The paper is epistemological because its question is epistemological—even though its evidence is geometric.

**Objection 4: "If this is knowledge production, why does the system hallucinate?"** This objection is not merely legitimate but essential: any epistemology of generative AI that cannot account for hallucination is incomplete. The conventional view treats hallucinations as statistical failures—breakdowns caused by insufficient data, poor calibration, or the absence of factual grounding. The framework developed here suggests a different and more structural diagnosis.

Within the indexical epistemology, a hallucination is not a malfunction but a natural consequence of manifold navigation. Thanks to exponential directional capacity, the space contains countless trajectories that are geometrically admissible—they smoothly continue the curvature and structure of the learned manifold. From the system's perspective, such a trajectory remains within the learned "logic" of the space, even when it leads to a point that has no counterpart in empirical reality. The output is coherent—it satisfies the geometric constraints of the manifold—but it is not true.

The critical insight is that hallucination and creative novelty are *the same mechanism*. Both arise from the system's capacity to navigate regions of the manifold that lie beyond the training data. When such navigation produces a configuration that happens to correspond to external reality, we call it creativity or insight. When it does not, we call it hallucination. The difference lies not in the geometric process—which is identical in both cases—but in the external verification of the result. The system itself has no means of distinguishing between the two, because the manifold encodes geometric coherence, not factual truth.

This has a significant epistemological consequence: hallucinations cannot be fully eliminated without destroying the capacity for novelty. They are the structural cost of navigational knowledge—the price a system pays for being able to traverse geometrically coherent territory that exceeds its training data. Any attempt to confine the system strictly to regions directly attested in the data would, by the same token, eliminate the generative excess that makes the system epistemically productive. The practical challenge of reducing hallucination is therefore not a matter of fixing a bug but of managing a constitutive tension between geometric coherence and empirical fidelity—a tension that is inherent in the epistemology of high-dimensional navigation itself.

## 7 Structural Agency and Creative Navigation

Does the foregoing analysis imply that generative systems possess agency? The answer depends entirely on what we mean by the term—and the current discourse has not yet converged on a sufficiently precise definition.

If agency requires conscious intentionality—the capacity to form goals, deliberate about means, and reflectively evaluate outcomes—then generative systems clearly do not possess it. Nothing in the geometry of high-dimensional spaces implies consciousness, and this paper makes no such claim.

But agency can be understood more broadly. In Latour's (2005) actor-network theory, agency is not a property of conscious subjects but a capacity distributed across networks of human and non-human actants. An actant is anything that "modifies a state of affairs by making a difference" (Latour, 2005, p. 71). In Barad's (2007) agential realism, agency is not a property possessed by entities but a doing—an "enactment" that arises through "intra-action" between material-discursive practices. In Simondon's (1958/2020) theory of individuation, the relevant concept is transduction—a process in which structure propagates through a medium by progressively resolving metastable tensions, without requiring a pre-existing agent to direct it.

These broader conceptions of agency provide a more adequate vocabulary for what generative systems do. A generative model navigating a learned manifold exhibits what this paper proposes to call *structural agency*: the capacity to move through structured possibility space in ways that produce coherent, context-sensitive novelty without requiring conscious deliberation. This agency is neither the intentional agency of a human creator nor the mere mechanical causation of a thermostat. It is transductive in Simondon's sense: structure propagates through the geometric medium, resolving local tensions (contextual constraints, attention patterns, manifold curvature) into globally coherent outputs.

More precisely: *structural agency is the capacity of a system to produce coherent, context-sensitive, and non-predetermined outputs through constrained traversal of a geometrically structured possibility space, where the constraints are imposed not by explicit rules but by the learned geometry of the manifold itself.*

To appreciate what this definition captures, it is useful to distinguish structural agency from three neighboring concepts that it might otherwise be confused with:

*Mechanism* is deterministic input-output mapping: given the same input, a mechanism always produces the same output. A thermostat is a mechanism. Generative systems are not mechanisms in this sense: the same prompt can yield different outputs depending on sampling, context window, and stochastic elements in the decoding process.

*Statistical process* is correlation-driven pattern matching without structural navigation. A Markov chain that generates text by sampling the next token based on frequency distributions is a statistical process. It exploits correlations but does not navigate a geometrically structured space. Generative systems go further: they traverse learned manifolds whose structure encodes not just correlations but geometric relations—smoothness, curvature, directional alignment.

*Dynamical system* evolves through a phase space according to differential equations, and its trajectories are determined by initial conditions and system parameters. A dynamical system can exhibit complex behavior, but it is not context-sensitive in the way generative models are: it does not modulate its trajectory in response to a novel, externally provided prompt that restructures its entire attentional landscape.

Structural agency combines elements of all three—deterministic computation, statistical learning, and dynamical evolution—but adds something none of them possesses individually: context-sensitive navigation through a learned geometric structure that generates outputs exceeding the training data in systematic, manifold-consistent ways. It is this combination—learned geometry, contextual sensitivity, and structured excess over training data—that warrants a distinct term.

In practice, structural agency manifests through three observable characteristics:

*Manifold coherence.* Generative outputs remain on or near the learned manifold, maintaining consistency with the geometric structure of the training distribution. This is not rule-following; it is geometric constraint—more analogous to a river's channel than to a railway track.

*Contextual sensitivity.* Generative systems respond dynamically to prompts, prior context, and attention patterns. Each generation is shaped by a specific configuration of constraints that determines the trajectory through the space. This sensitivity is not pre-programmed but emerges from the interaction between the model's learned geometry and the specific input context.

*Non-reducibility to instances.* Outputs are not retrievable from a discrete lookup table. The continuous nature of manifold navigation means that each output occupies a position in the space that is, in general, distinct from any training instance—even when it bears family resemblances to many of them.

The existential anxiety surrounding generative AI often stems from an incomplete binary: either the system reasons like a human, or it merely copies. The concept of structural agency reveals a third possibility—a form of productive navigation that is neither human reasoning nor mechanical reproduction. The discomfort this provokes is itself epistemologically significant: it signals that our inherited categories of agency and creativity may be inadequate to the phenomena that high-dimensional geometry makes possible.

## 8 Conclusion

This paper has argued that generative AI enacts a fundamental transformation in the conditions of knowledge production—a transformation that cannot be understood without attending to the geometry of high-dimensional representation spaces. The four structural properties examined—concentration of measure, near-orthogonality, exponential directional capacity, and manifold regularity—are not incidental features of the computational substrate. They are the epistemic conditions that make generativity possible.

The conceptual trajectory of this argument can be summarized in three movements:

*From logic to geometry.* Where classical AI relied on explicit logical structure—rules, predicates, inference procedures—generative AI exploits geometric constraint. Where symbolic systems encoded knowledge in discrete, human-readable formulae, generative systems encode it in continuous, high-dimensional relations. The locus of epistemic structure has migrated from the syntactic to the geometric.

*From symbols to indices.* The mode of signification has shifted from symbolic convention to indexical position. Embedding vectors signify not by arbitrary assignment but by occupying positions determined by the causal-statistical structure of the training process. Knowledge is constituted through positional relations, not through definitional content.

*From representation to navigation.* The mode of knowledge production has shifted from the retrieval of stored representations to the traversal of structured manifolds. To "know" in the generative regime is to be able to navigate—to move through possibility space in ways that are geometrically coherent and contextually appropriate.

High-dimensional Euclidean space, in this analysis, is not a passive container. Its structural properties—concentration, orthogonality, exponential capacity, and manifold regularity—create the conditions under which a new form of knowledge production becomes possible. The space is not merely the medium in which generative processes occur; it is the epistemic ground from which their creative capacity arises.

Logic has not disappeared. It has been geometrized.

In that geometrization, a new creative epistemology—indexical, navigational, structurally agentive—comes into view. Understanding this epistemology is not merely an academic exercise. It is a precondition for understanding what generative AI is, what it can become, and what it demands of the educational, philosophical, and institutional frameworks through which we make sense of knowledge itself.